\documentclass[11pt]{article}
\usepackage[final]{acl}

\usepackage{microtype}

\usepackage[english,bidi=default]{babel} 
\babelfont{rm}{TeXGyreTermesX} 

\usepackage{graphicx}

\usepackage[table]{xcolor}

\title{Using Correspondence Patterns to Identify Irregular Words in Cog\-nate sets Through Leave-One-Out Validation} 

\author{
 \textbf{Frederic Blum\textsuperscript{1,2}} \quad\quad
 \textbf{Johann-Mattis List\textsuperscript{1,2}}
\\
 \textsuperscript{1}Department of Linguistic and Cultural Evolution, \\ Max Planck Institute for Evolutionary Anthropology, Leipzig, Germany\\
 \textsuperscript{2}Chair for Multilingual Computerlinguistics, University of Passau, Passau, Germany
\\
 \small{
   \textbf{Correspondence:} \href{mailto:frederic\_blum@eva.mpg.de}{frederic\_blum@eva.mpg.de}
 }
}

\begin{document}
\maketitle
\begin{abstract}
Regular sound correspondences constitute the principal evidence in historical language comparison.
Despite the heuristic focus on regularity, it is often more an intuitive judgement than a quantified evaluation, and irregularity is more common than expected from the Neogrammarian model. Given the recent progress of computational methods in historical linguistics and the increased availability of standardized lexical data, we are now able to improve our workflows and provide such a quantitative evaluation.
Here, we present the balanced average recurrence of correspondence patterns as a new measure of regularity. We also present a new computational method that uses this measure to identify cognate sets that lack regularity with respect to their correspondence patterns. We validate the method through two experiments, using simulated and real data. In the experiments, we employ leave-one-out validation to measure the regularity of cognate sets in which one word form has been replaced by an irregular one, checking how well our method identifies the forms causing the irregularity.
Our method achieves an overall accuracy of 85\% with the datasets based on real data. We also show the benefits of working with subsamples of large datasets and how increasing irregularity in the data influences our results.
Reflecting on the broader potential of our new regularity measure and the irregular cognate identification method based on it, we conclude that they could play an important role in improving the quality of existing and future datasets in computer-assisted language comparison. 
\end{abstract}

\section{Introduction}
Recurring correspondence patterns are the cornerstone of the comparative method and historical linguistics \cite{Anttila1972, Fox1995, Durie1996}. They are built upon the definition of regular sound change as a process that occurs across the whole lexicon in a specified set of contexts \cite[XIII]{Osthoff1878}. However, regularity is often more an intuitive judgement than a quantified evaluation, and irregularity is more common than expected from the Neogrammarian model of language change \cite{Ross1996, Nichols1996, Blust2022}.

Previous computational approaches have used instances of regular sound change to infer the correspondence patterns of a comparative wordlist \cite{List2019} or to compare the regularity of datasets before and after applying methods that alter the correspondence patterns in the data \cite{blum2023trimming}. Implicitly, they also form the basis for automated approaches of reconstructing proto-languages \cite{List2023fuzzy}. Yet, the full utility that regularity provides is rarely exploited.

Here, we present a new measure that evaluates the regularity of correspondence patterns in a standardized comparative wordlist. By computing the balanced average recurrence of correspondence patterns, we establish a score that can be used to identify cognate sets without such recurrent patterns. Through different normalizations, we can also compare this measure across datasets. We illustrate how this workflow can be used with a new method that detects erroneous cognate judgements in comparative wordlists. Our approach measures the regularity of cognate sets and reports on the cases with a low score. We can then implement a leave-one-out validation and iterate through each cognate set to identify those word forms whose deletion improves the regularity of the cognate set.

\section{Background}
Sound change has been recognized as an overwhelmingly regular process \cite{Osthoff1878}. If a sound change occurs in the lexicon of a given language at a certain point in time, it usually affects all of the words that occur in a particular phonetic environment, leaving almost no exceptions \cite{Campbell1996}. Based on this regularity, scholars have developed methods for the reconstruction of proto-languages which are not attested in written sources, but that can be inferred through the systematic comparison of related languages. Those reconstructed languages can then be used for further comparisons to establish genealogical links between different language families.

Regularity of sound change is the core principle of linguistic reconstruction. Linguists identify recurring correspondence patterns through the annotation and alignment of cognate sets, that is, sets of related words in a language family \cite{Anttila1972, Fox1995}. Those recurring correspondence patterns build the basis for the reconstruction of proto-phonemes. Due to the regularity of those patterns, they can also be used for predicting reflexes in languages where we have no explicit evidence of the lexical form. Such approaches have been used for targeted fieldwork \cite{Bodt2021} or testing hypotheses of genealogical relationships between language families \cite{Blum2024d}. The regularity of correspondence patterns also provides the arguments to distinguish chance similarity or borrowings from true genealogical descent.

The regularity of sound change doesn't describe dogmatic laws, but rather the definition of a \textit{specific type} of sound change that differs from irregular forms of change such as analogy and borrowings \cite{Hoenigswald1978, Labov1981}. Those are considered to be of minor importance to the overall regularity, which keeps forming the basis for linguistic reconstructions \cite{Ross1996, Campbell1996}. The cognate annotation for related languages based on the regularity principle is still considered the state-of-the-art for historical linguistics. It's applications in the 21st century include the manual reconstruction of proto-phonemes \cite{Zariquiey2026}, the relative dating of sound changes \cite{Fries2024}, the prediction of cognate reflexes \cite{Blum2024d}, and the preparation of comparative wordlists for various computer-assisted methods \cite{Wu2020, Blum2024}.

Within computer-assisted workflows, comparative wordlists annotated for cognate sets are also used in phylogenetic studies that infer the temporal dimension of the diversification of language families, as well as their internal classification \cite{Greenhill2020}. Comparative wordlists can be built in a standardized way using the Cross-Linguistic Data Formats (CLDF, \citet{Forkel2018}). The standardization includes linking languages to Glottolog \cite{Glottolog} and phonetic segments to the Cross-Linguistic Transcription Systems \cite{CLTS}. The standards established for LexiBank datasets \cite{List2022, Blum2025b} also allow for a transparent annotation of cognate sets. For example, such cognate-annotated wordlists have been used to study the history of Austronesian \cite{King2024}, Bantu \cite{Grollemund2015}, Indo-European \cite{Heggarty2023}, Mixtecan \cite{Auderset2023}, Sino-Tibetan \cite{Sagart2019}, Uto-Aztecan \cite{Greenhill2023}, and other language families. However, no computational methods exist to evaluate the presented cognate coding, and more often than not, the data is taken as granted by reviewers and readers. This is problematic in many ways, since different interpretations of the same data often co-exist \cite{Anderson2025, Kassian2025}, and the data should be considered a fundamental part of any statistical model \cite{Mcelreath2020}. However, it has not really been defined how the regularity can be measured, and how individual cognate sets can be evaluated with computer-assisted methods.

\section{Materials and Methods}
\subsection{Cognate-Coded Comparative Wordlists}
\begin{figure*}[t!]
    \centering
    \includegraphics[width=\linewidth]{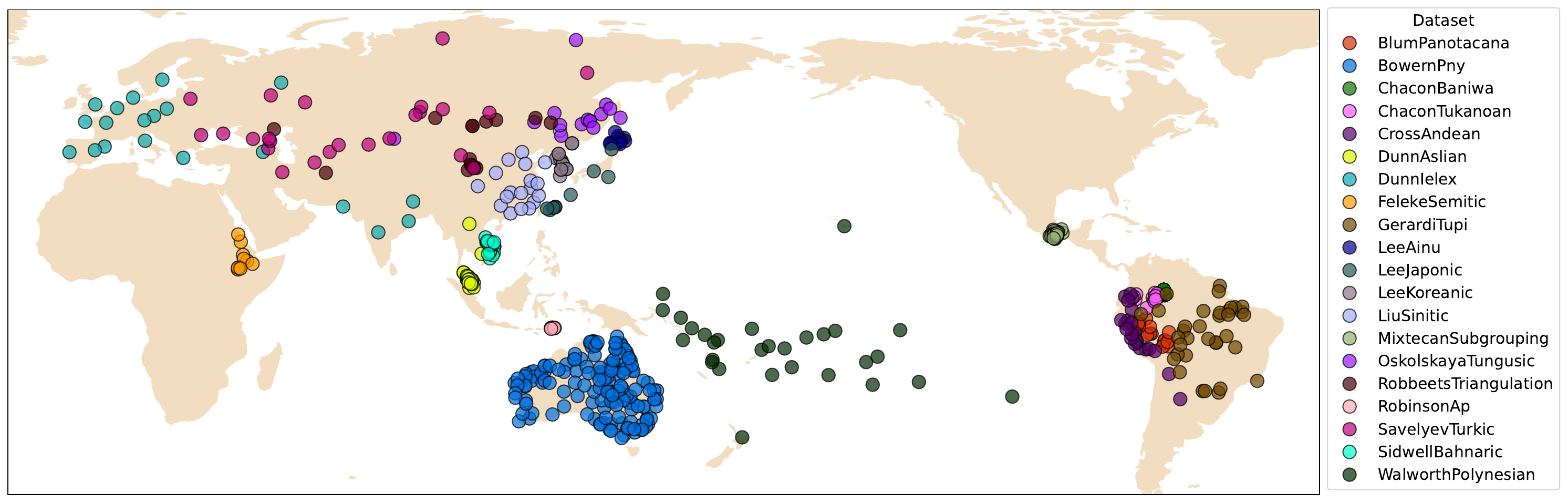}
    \caption{Map of all languages in the sample coloured per dataset.}
    \label{fig:map}
\end{figure*}

For our study, we use 20 datasets from LexiBank v2.1 \cite{Blum2025b}. The datasets are available in the Cross-Linguistic Data Formats \cite{Forkel2018}, and are mapped to Glottolog \cite{Glottolog}, Concepticon \cite{Concepticon}, and CLTS \cite{CLTS}. The sample is part of the CogCore subset of LexiBank, which only includes datasets with manually annotated cognacy. We analyse the majority language family of each dataset and preprocess the data via LexiBench \cite{Häuser2025}, which provides a number of modifiable thresholds. We have only included datasets with at least ten languages from the same family and a minimum number of 140 concepts mapped to Concepticon \cite{Concepticon}. We present the resulting 576 languages from 19 language families in Figure~\ref{fig:map}, and the number of languages and cognate sets per dataset in Table~\ref{tab:data}.

One problem for the development of methods that exploit the regularity of correspondence patterns is the lack of datasets with manual alignments. The only datasets in our sample that include such manual alignments for the whole comparative wordlist are \texttt{CrossAndean} \cite{Blum2023-quechua} and \texttt{BlumPanotacana} \cite{Blum2024}. Tests on large benchmark datasets have shown that multiple phonetic alignment algorithms work rather well, achieving about 88\% of identical columns on considerably divergent cognate sets \citep{List2014d}. The benchmarks on which these algorithms were tested, however, did not reflect the kind of data that we observe in comparative wordlists, but rather consisted of hand-selected cognate sets from closely related dialect varieties. As a result, we expect that automated alignment methods serve as an additional source of irregularities in the data, since the data in comparative wordlists often comes from languages that not as closely related as the sampled dialect varieties on which the algorithms have been tested.

\begin{table*}[t!]
    \centering
    \small
    \begin{tabular}{|l|l|l|c|c|c|c|}
        \hline
         \rowcolor{gray!35}Dataset & Family & Source & Languages & Concepts & Cog\-nates \\ \hline\hline
            BlumPanotacana & Pano-Tacanan & \citet{Blum2024} & 17 & 419 & 487 \\
            BowernPny & Pama-Nyungan & \citet{bowernpny} & 175 & 238 & 2171 \\
            ChaconBaniwa & Arawakan &  \citet{chaconbaniwa} & 14 & 217 & 263 \\
            ChaconTukanoan & Tucanoan & \citet{chacontukanoan} & 16 & 141 & 141 \\
            CrossAndean & Quechuan & \citet{Blum2023-quechua} & 34 & 150 & 245 \\
            DunnAslian & Austroasiatic &  \citet{dunnaslian} & 31 & 145 & 369 \\
            DunnIelex & Indo-European & \citet{dunnielex} & 20 & 205 & 485 \\
            FelekeSemitic & Afro-Asiatic & \citet{Feleke2021}  & 21 & 150 & 273 \\
            GerardiTupi & Tupian & \citet{gerarditupi} & 37 & 243 & 395 \\
            LeeAinu & Ainu & \citet{leeainu} & 19 & 195 & 289 \\
            LeeJaponic & Japonic & \citet{leejaponic}  & 56 & 197 & 382 \\
            LeeKoreanic & Koreanic & \citet{leekoreanic}  & 14 & 206 & 230 \\
            LiuSinitic & Sino-Tibetan & \citet{liusinitic} & 19 & 202 & 306 \\
            MixtecanSubgrouping & Otomanguean & \citet{mixtecansubgrouping}  & 84 & 224 & 436 \\
            OskolskayaTungusic & Tungusic & \citet{Oskolskaya2022} & 21 & 254 & 498 \\
            RobbeetsTriangulation & Mongolic &  \citet{robbeetstriangulation} & 15 & 253 & 415 \\
            RobinsonAp & Timor-Alor-Pantar & \citet{robinsonap}  & 13 & 217 & 256 \\
            SavelyevTurkic & Turkic & \citet{savelyevturkic}  & 31 & 254 & 508 \\
            SidwellVahnaric & Austroasiatic &  \citet{sidwellbahnaric} & 24 & 199 & 402 \\
            Walworthpolynesian & Austronesian & \citet{walworthpolynesian} & 31 & 210 & 511 \\
        \hline
    \end{tabular}
    \caption{The datasets used for the experiment, including their number of languages, concepts, and cognate sets.}
    \label{tab:data}
\end{table*}

\subsection{Measuring Regularity}
Following the major findings made by linguists in the beginning of the 19th century, we know that sound change proceeds in an overwhelmingly regular manner \cite{Osthoff1878, Hoenigswald1978}. So far, however, quantitative measures for this particular regularity are lacking. A valid regularity measure for comparative wordlists must account for three domains, in which different notions of regularity play an important role. First, there are individual correspondence patterns. These are abstract representations of the columns (sites) of all alignments of the cognate sets in a given comparative wordlist. Based on their mutual compatibility, the individual alignment sites can be grouped together into correspondence patterns \cite{List2019}. The alignment sites that constitute a correspondence pattern can be counted directly, and we assume that regularity of individual patterns can be represented through frequency. Second, there are the cognate sets, which are groups of words that were found to be related. The segments within each cognate set can be phonetically aligned, creating the alignment sites mentioned above. For each cognate set, we would like to measure how regular the combination of individual correspondence patterns underlying a particular alignment is compared to other alignments in the same dataset. Third, there is the dataset as a whole, which should display a certain balance, both in the number of regular cognate sets and the number of regular correspondence patterns.

\begin{figure*}[t!]
    \centering
    \includegraphics[width=0.95\linewidth]{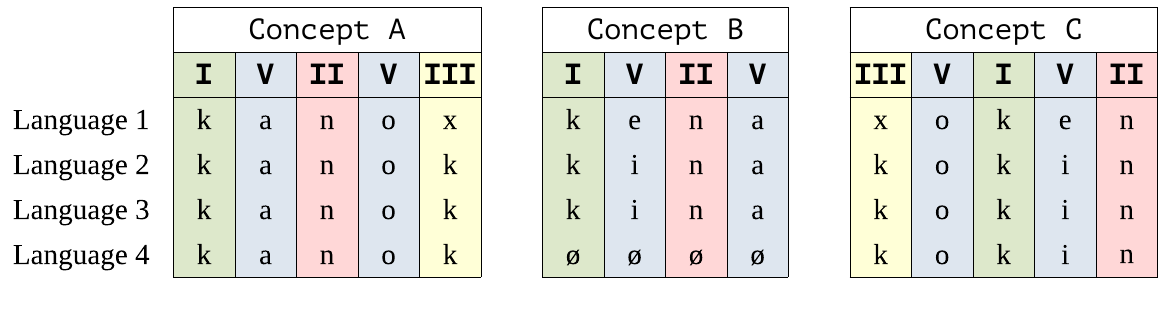}
    \caption{Artificial example for three cognate sets across four languages with their aligned segments. The columns are numbered and coloured according to their correspondence patterns. The vowel patterns are not further distinguished since all examples in the text refer to the patterns I-III.}
    \label{fig:alignments}
\end{figure*}

The grouping of alignment sites into correspondence patterns is shown in Figure~\ref{fig:alignments}. Here, we see the cognates from four artificial languages for three different semantic concepts. The segments in all cognate sets are aligned phonetically. The individual columns (alignment sites) are then grouped into correspondence patterns based on their compatibility. For example, pattern I has the same reflex /k/ in all languages. In the case of concept B, the missing value for Language 4 is imputed based on the compatibility of the individual sites \cite{List2019}. Similarly to pattern I, pattern II features /n/ in all languages, with an imputed value in Language 4 for Concept B. Pattern III is distinguished from pattern I based on the contrastive reflex in Language 1, where we have /x/ (III) instead of /k/ (I).

For our analysis, we assume that cognates, alignments, and correspondence patterns have been inferred or computed in advance. Correspondence patterns are represented as a tuple of the length of the language varieties in a given wordlist. For each language, either the sound that would be expected in a regular alignment site is listed, or a specific symbol for missing data, in case the pattern cannot be fully resolved. Patterns are linked to individual alignment sites. In turn, the alignment sites can be linked to one or more patterns. Since the individual alignment sites may show varying amounts of missing reflex sounds in individual languages -- given that cognate sets are rarely present in all languages of a family -- they may show differential \emph{compatibility} with alternative correspondence patterns (see \citealt{List2019} for details).

Our workflow for measuring the regularity of correspondence patterns in comparative wordlists consists of three steps. The starting point is an individual cognate set, represented as a phonetic alignment. This was illustrated in Figure~\ref{fig:alignments}. In the first step, we iterate over all \emph{sites} in the alignment and identify all its compatible correspondence patterns. For each site, we select the pattern that is compatible with the highest number of individual alignment sites. This number reflects the \emph{recurrence across sites} of the pattern. Calculating this recurrence for all sites of a given alignment yields a list consisting of the individual site recurrence scores, one for each column. To quantify the regularity of a given cognate set, we first log-transform all individual recurrence scores and then compute the mean. This transformation helps to balance out potentially skewed distributions that may result in those cases where one pattern has a high recurrence score while all other patterns occur only once. If we take, for example, an alignment of four sites, with a site recurrence of \texttt{[1 1 1 15]}, it would yield a higher absolute mean (4.5) than an alignment with a site recurrence of \texttt{[2 2 2 2]} (2). Taking the mean of the log-values will result in ranking the second set higher instead (0.67 vs 0.69). We can then take the exponential of this mean to arrive at a log-balanced average recurrence of individual sites within the cognate set (1.97 vs 1.99).

\begin{figure*}[t!]
    \centering
    \includegraphics[width=\linewidth]{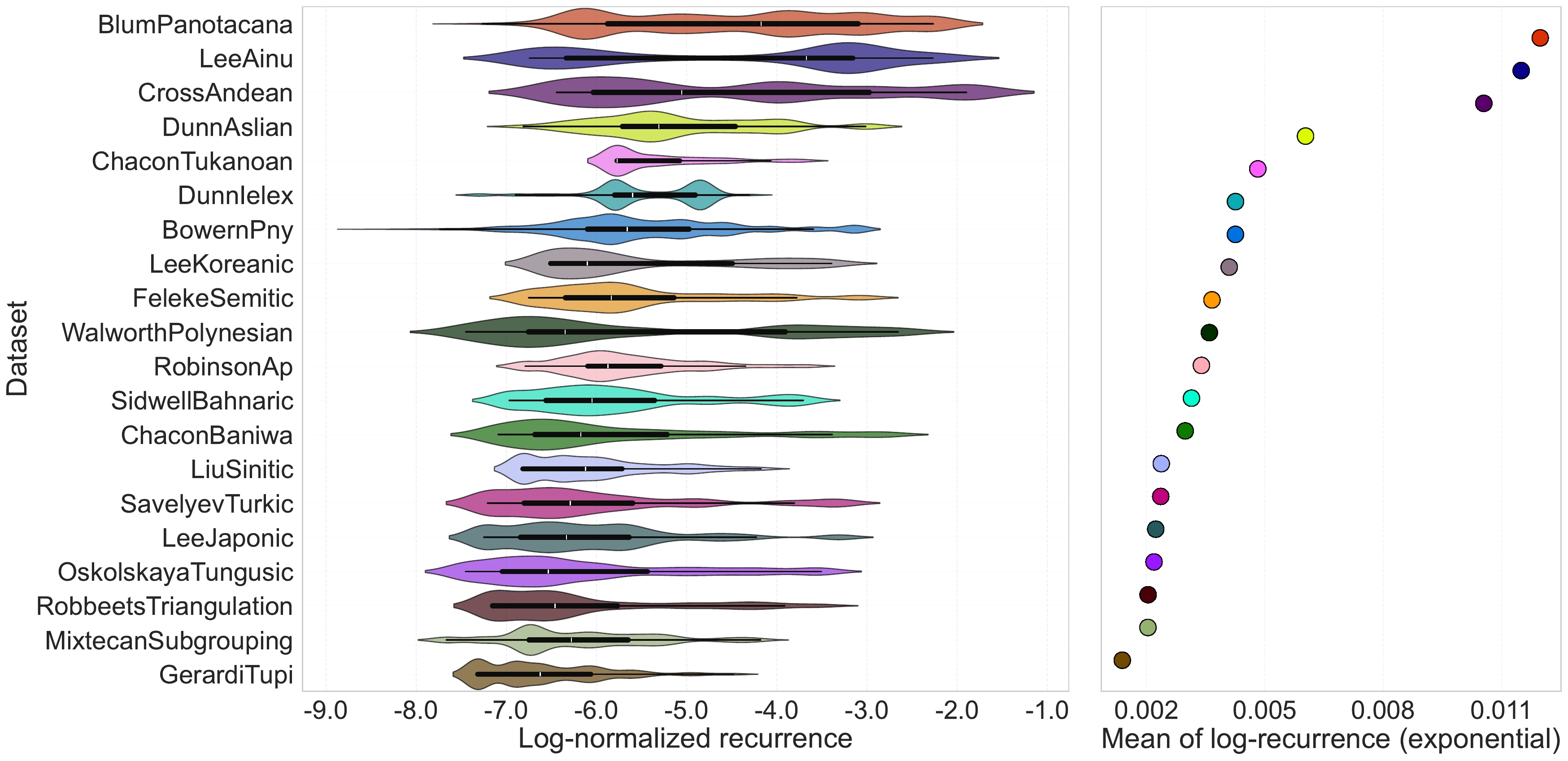}
    \caption{Overall regularity of the 20 datasets in the sample (y-axis) as indicated by measuring the average recurrence of sites in two ways. The left subplot shows the normalized and log-transformed recurrence of each site (x-axis). The right subplot shows the exponential of the mean of that log-transformed recurrence. This score can be interpreted as the balanced average pattern recurrence of an alignment site within each dataset.}
    \label{fig:overall}
\end{figure*}

The third domain, regularity in the overall dataset, comes with an additional complication. Datasets differ in size with respect to the number of languages, concepts, and cognate sets they contain. In order to compare for regularity across datasets, we need to normalize the data. The first step here is to normalize the recurrence of each site through the total number of sites in the data. The normalized value is then log-transformed. In order to reach a regularity score that is comparable across datasets, we can take the exponential of the mean of the logarithm of the normalized values. This score describes the balanced average recurrence of a site within the dataset, and can be compared across datasets. The steps to reach this score can be summarized as follows: (1)~counting the recurrence of alignment sites in the data, (2)~normalize the count by the total number of sites, (3)~log-transform the recurrence, and (4)~take the exponential of mean of the log-recurrence (see Figure~\ref{fig:overall}).

According to our new regularity measure, we can observe in Figure~\ref{fig:overall} that three datasets in our sample are far more regular than the others: \texttt{CrossAndean}, \texttt{BlumPanotacana}, and \texttt{LeeAinu}. What sets the first two apart qualitatively is that they are the only datasets for which we have manual alignments of the cognate sets. The fact that they score higher with respect to the recurrence of correspondence patterns is thus at least partially a display of the careful annotation of the data. Another important factor seems to be that in the first and the third dataset, the languages are very closely related. In those cases, irregularities are less likely to arise. With increasing distance between the languages, the irregularity in cognate sets tends to increase, since there is more time for sporadic and irregular changes to occur. 

The correspondence pattern measure suffers from another particular problem: with increasing language numbers, the methods will produce fewer and fewer regular correspondence patterns, since each language that is added to the selection may contain individual irregularities that then sum up to global irregularity. We will showcase this issue with simulated data. Another reason for the problem of apparently increasing irregularity is that correspondence patterns are not identical with proto-sounds, since individual regular sound changes may lead to a situation where one original proto-sound diversifies into several correspondence patterns. As a result, adding more languages to the comparison will almost always increase the number of correspondence patterns, due to individual sound changes that happen across branches.

\subsection{Identifying Irregular Words}
We confirm the utility of our measure through a new computational method that detects irregular words in cognate sets. For this purpose, we set up two experiments. In both experiments, we replace a single word form from an existing cognate set with a randomly generated irregular word. While the first experiment uses simulated regular data, the second experiment injects irregularities into real comparative wordlists (see \citealt{Dessimoz2008} for this technique in the context of lateral gene transfer in biology and \citealt{List2015d} for an application in the context of language contact in linguistics). In both experiments, we apply our new method to identify the most irregular form in a cognate set. In the experiment with the simulated data we show how pre-existing irregularity in correspondence patterns affects the accuracy of our method, and in the second we show how the method can successfully identify artificial forms that were injected into real cognate sets.

For the replacements, we first create a dictionary for each language that stores all its consonants and vowels. Then, we replace a randomly chosen word form in 20\% of the cognate sets of each dataset. We replace each segment with a random entry from the dictionary for that language that is different to the original one, ensuring a form that has different segments than the original word form.

The basic workflow of our identification method is a leave-one-out validation. Within each of the cognate sets in the data where we have replaced a word form, we iterate through all its alignment sites and run the leave-one-out validation to compute the regularity. One word form at a time, we set the reflexes within the alignment to /ø/ and verify whether the altered alignment site can be matched to a compatible correspondence pattern in the data with higher recurrence than previously. By repeating this iteration for all alignment sites, we can identify the word form whose exclusion leads to the highest gain in regularity for that cognate set, measured as the average log-recurrence of its alignment sites. This identification directly relates to the second domain of regularity described above, which compares the average log-recurrence of sites between two cognate sets. We then verify whether the chosen form is the one which we have injected into the cognate set to evaluate the accuracy of our method. Both the replacement workflow and the evaluation are identical in both experiments.

\subsubsection{Simulating Irregular Data}
We use simulated data to show how pre-existing irregularity within comparative wordlists affects our method of identifying irregular word forms in cognate sets. For this purpose, we run code that simulates proto-forms for 200 concepts, based on a phoneme inventory including four vowels and ten consonants, and a CVCV syllable structure. For each of the ten simulated daughter languages, we simulate 0 to 2 regular phoneme mergers, each involving the random selection of two phones from the vowel or consonant inventory. This ensures that we don't have any form identity in the cognate sets of the simulated data. The correspondence patterns are still fully regular, since we have the control over the data-generating process and apply all changes in all relevant instances. While neither the full regularity nor the sampled sound changes are a realistic scenario for real linguistic data, it allows us to verify the method in a controlled environment that involves an initial level of complexity. One such simulated cognate set is presented in Table~\ref{tab:simulated}.

\begin{table}[t]
    \centering
    \resizebox{0.95\linewidth}{!}{%
    \begin{tabular}{|c| c c c c |l|} \hline
        \rowcolor{gray!35}
        \textbf{Sample} & \multicolumn{4}{c}{\textbf{Form}} & \textbf{Sound Changes} \\ \hline 
        A & r & a & l & a & */e/, */a/ > /a/ \\ \hline 
        B & r & a & l & e & \\ \hline 
        C & r & e & l & e & */e/, */a/ > /e/ \\ \hline 
        D & r & a & k & e & */k/, */l/ > /k/ \\ \hline 
        E & r & a & l & e & \\ \hline 
        F & r & a & l & e & \\ \hline 
        G & j & i & l & e & */i/, */a/ > /i/; */j/, */r/ > /j/ \\ \hline 
        H & r & a & l & i & */e/, */i/ > /i/ \\ \hline 
        I & w & a & l & i & */e/, */i/ > /i/; */w/, */r/ > /w/ \\ \hline 
        J & r & a & l & e & \\ \hline 
    \end{tabular}}
    \caption{Phonetic alignments for word forms across ten simulated languages with random phoneme mergers for some languages.}
    \label{tab:simulated}
\end{table}

For the experiment, we run ten different settings where we randomly replace between 5\% and 50\% of phones in the data. This leads to an artificial decrease in the overall regularity of the dataset through injecting random irregularities into the cognate sets, giving us a first indication how pre-existing irregularity in a comparative wordlist affects the inference of correspondence patterns.

\begin{figure*}[t]
    \centering
    \includegraphics[width=0.97\linewidth]{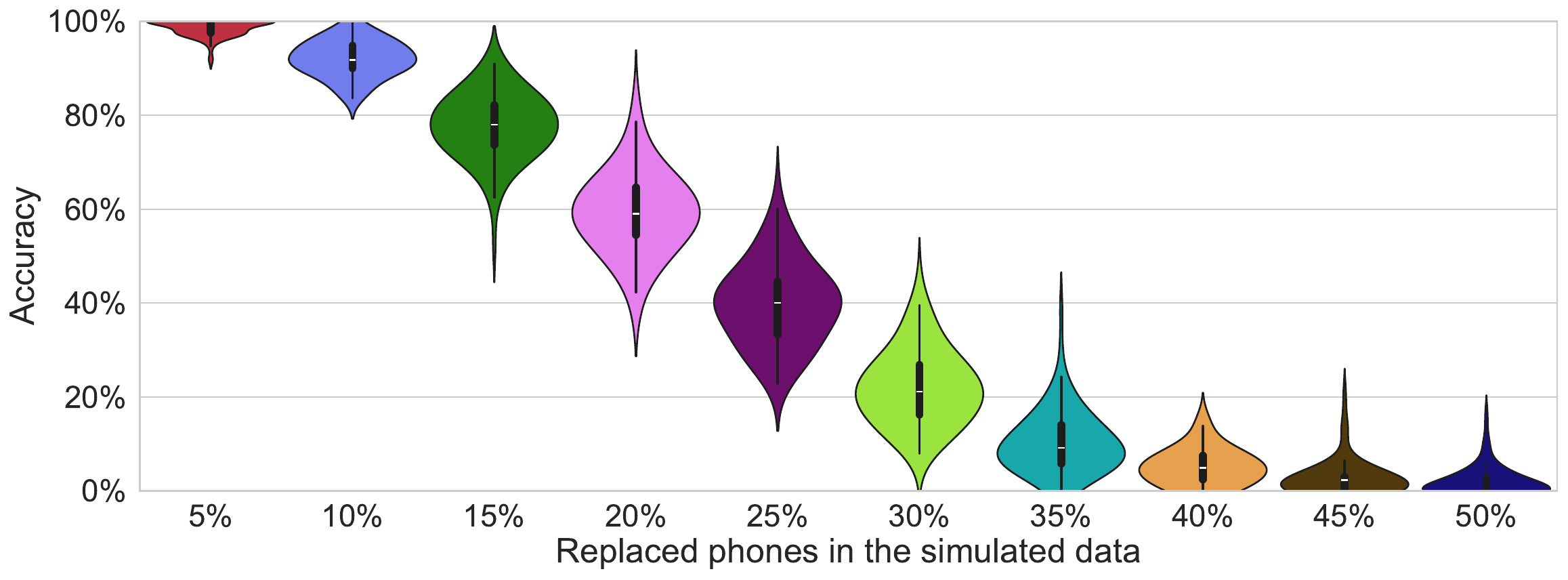}
    \caption{Accuracy for the leave-one-out validation with simulated regular data (y-axis) and random replacement of phones (x-axis) to simulate different levels of irregularity in comparative wordlists.}
    \label{fig:simulated}
\end{figure*}

\subsubsection{Injecting Artificial into Real Data}
\label{method: loo}
The second experiment uses the 20 comparative wordlists presented in Table~\ref{tab:data}. All datasets include manually annotated cognate sets provided by the original study authors. The correspondence patterns are inferred using a subsample of languages from each dataset. We distinguish two settings: one that samples five languages, and one that samples ten. This step ensures comparability across datasets, which widely differ in the number of languages they feature. In order to compare the influence of this subsampling, we implement 100 runs for each dataset in each experimental condition and compare the distribution of the accuracy. In total, this results in 4,000 experimental trials.

This method has one big challenge: as we have already seen in Figure~\ref{fig:overall}, the datasets are not perfectly regular to begin with. The contrast with the simulated data shows that the real world of sound change is messy, and we do not expect our method to work perfectly. But once the task is defined, we can work on improving the methods and seek ways to counterbalance the factors contributing to the irregularity in the datasets. We will come back to this topic in the conclusions.

\subsection{Implementation}
We create Python code with a new dedicated package that provides the functionality for regularity computation. Cog\-nate sets are already provided along with the individual cognate sets in our data sample. Where missing, phonetic alignments are computed with the help of the SCA algorithm \citep{List2014d}, implemented in LingPy \citep{LingPy}. Correspondence patterns are inferred with the help of the CoPaR algorithm \citep{List2019}, implemented in LingRex \citep{LingRex}. All code and data needed to reproduce the results presented here are linked in the Data Availability Statement.

\section{Results}
\subsection{Experiment with Simulated Data}
In the first experiment we analyze the simulated data. This experiment not only shows how our method works in principle, but also how irregularity in the data affects its accuracy. The results for this experiment are presented in Figure~\ref{fig:simulated}.

The hypothesis that the accuracy of the method decreases with increasing irregularity in the data holds. Even for the otherwise perfectly regular data, accuracy drops to $\sim$90\% when replacing 10\% of phones. Starting at around 40\% of replaced phones, our method drops below 10\% accuracy on average. This indicates that the success of our method heavily depends on the initial regularity of the dataset. The less regular our dataset is in general, the more difficult it becomes to correctly identify the irregular forms. This simulation shows the complexity of the task once irregular correspondence patterns, expected to be abundant in the real data, creep in. But it also shows that our new measure can be applied successfully within computational methods that compare the regularity of competing alignment sites.

\subsection{Replacing Sounds in Real Data}
\begin{figure*}[t]
    \centering
    \includegraphics[width=\linewidth]{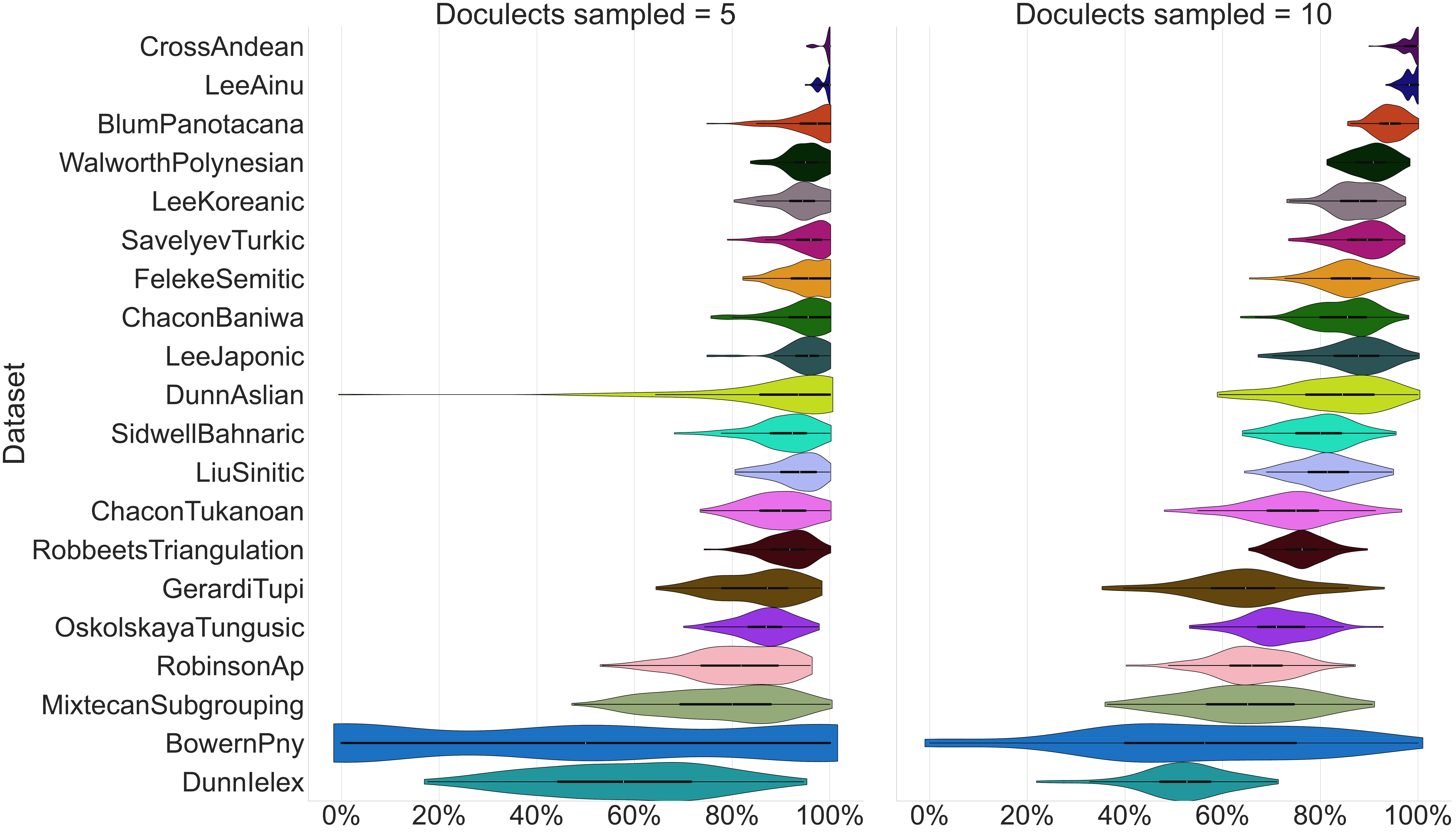}
    \caption{Distribution of experiment results for all datasets, split into the 5-doculects and 10-doculects sample settings. The boxplot presents the 50\% distribution of all results.}
    \label{fig:results}
\end{figure*}

In the second experiment, we use 100 random subsample of five and ten languages for all 20 datasets. In each subsample, we replace one lexical word form in 20\% of the cognate sets and verify whether our method correctly identifies the replaced form. The results for the experiment are presented in Figure~\ref{fig:results}. Our method reaches an overall accuracy of 85\%, with the dataset means varying between 52\% and 99\%. This shows that the success of the method is largely dependent on the individual dataset. The results also show that the individual runs show a great variability in accuracy, indicating that the language sample and choice of cognate sets are important factors.

For all datasets, sampling only five languages instead of ten provides a considerable improvement in  accuracy, confirming the potentially devastating impact of accumulating irregularities. The average accuracy for the five-language sample is 87.8\% ($\sigma$20\%), dropping to 79\% ($\sigma$17\%) for the 10-language sample. This confirms that the regularity in the data is higher with fewer languages sampled, most likely because there are fewer individual reflexes disturbing the regularity of the correspondence patterns. The more languages we have in the data, the more possibilities for a single irregular site that disturbs the patterns arise. From this perspective, every additional language is another potential source for irregularities in alignment sites. We have seen a similar phenomenon with the simulated data. The more irregularity we find within individual cognate sets, the more difficult it is to identify the injected irregular form.

The same datasets top the accuracy that also showed the highest regularity in Figure~\ref{fig:overall} to begin with: \texttt{CrossAndean} (99\%), \texttt{LeeAinu} (99\%), and \texttt{BlumPanotacana} (95\%). At the other end of the scale, there are two datasets that stand out with respect to the lowest amount of accuracy in the experiment: \texttt{BowernPny} (53\%) and \texttt{DunnIelex} (55\%). They are the only datasets with an overall average below 70\%. Arguably, both include some of the most distantly related individual languages in our sample. \texttt{BowernPny} is also by far the largest dataset in the sample and represents one of the language families where automated approaches to language affiliation fail to achieve a large accuracy as well \cite{Blum2025c}. But the exact reasons why they perform worse 
cannot be answered 
here. 

\section{Conclusions}
The regularity of language change is a complex phenomenon to which computer-assisted language comparison contributes a novel, quantitative perspective. We have presented a new measure that evaluates the average recurrence of correspondence patterns in standardized comparative wordlists. In contrast to previous implementations, this measure balances out several of the complexities of correspondence patterns -- like their skewed distribution and the number of languages in the comparative wordlist -- and is comparable across datasets. We have implemented this measure within a new computational method that successfully identifies irregular cognate sets and identifies those words that should potentially be excluded from the set. Two experiments using simulated and real datasets confirm the utility of the new measure and highlight different aspects of irregularity in comparative wordlists.

In the first experiment, using simulated data, our method reaches over 90\% accuracy when randomly replacing up to 10\% of the phones from the correspondence patterns. With increasing irregularities, the accuracy of our method decreases until dropping below 10\% at 40\% of replaced phones. The situation is different with the real datasets, where we find substantial differences between individual datasets. Those differences are influenced by the overall regularity in the dataset, as well as by the degree of relatedness between the languages in the data. On average, our method reaches 87.5\% of accuracy when sampling five languages. The accuracy is lower when more languages are sampled.

There are many sources for irregularity, ranging from annotation errors, inconsistencies in the alignment algorithm, up to borrowings, or sporadic and irregular forms of sound change. Having identified the challenges, we can now focus at resolving them. First, we need to increase the data quality. While phonetic standardization through CLDF is a great first step, we need datasets that have also been manually aligned by experts. This will lead to an immediate increase in the regularity of the data, since no irregular patterns are introduced through automated alignment. Manual alignments could also lead to improved cognates, increasing in turn the regularity of alignment sites in the data, and resulting in better proto-language reconstructions.

We have also shown that sample size and language choice matter. The large deviations from the mean in the different experiment runs indicate that the accuracy heavily depends on the fact which languages and cognate sets have been sampled. The results also change with respect to how many languages are sampled, and the role of sample size in this kind of task will likely be a focal point of interest in the future. Through a more detailed analysis of this aspect, it should be possible to identify the individual languages which lead to an increased irregularity. These are likely the languages which represent key information for the history of the language family, since they might feature retentions and innovations not observed in other languages. By understanding which languages are especially complex in relation to the correspondence patterns can thus be of large benefit to historical linguists and provide important insights into the history of individual language families.

Our measure also provides a basis for additional applications. Regularity of cognate sets can and perhaps should be a commonly used tool for evaluating data in computer-assisted language comparison. For example, it could be used within computational methods that evaluate the cognate coding used for phylogenetic analysis and improve the cognate coding within such datasets. It could also point readers or reviewers to possibly problematic cognate judgements which require further explanation. It can also be used to help the traditional reconstruction of language families by evaluating cognate judgements in the same way. We present such an application as part of our package, based on the leave-one-out comparison from the experiments. Another pathway can be the extension of the analysis to correspondence patterns involving proto-phonemes, and to identify those cognate sets that deviate from the proposed patterns.

\section*{Limitations}
The main limitation is the availability of manually curated data for alignments, since the alignment of cognate sets is the crucial step previous to the inference of correspondence patterns. While there exist several datasets with manually annotated cognacy, there is only a handful of datasets with manual alignments. Another limitation is that the candidate selection of our method only works if there is exactly one entry causing the irregularity. While our method can always identify the irregular cognate sets, the leave-one-out methodology can only capture improvements if leaving out a single word forms leads to an increased regularity of a cognate set. If there are two irregular segments within the alignment, then no such improvement will be observed. In those cases, our method will be able to highlight the low regularity of the cognate set, but not be able to identify the word form(s) causing it.

\section*{Data Availability Statement}
All code that has been used for this study is curated on Codeberg (\url{https://codeberg.org/calc/lingreg}, Version 0.1), containing a Python package and instructions how to run the code.

\section*{Funding}
This research was supported by the Max Planck Society Research Grant CALC3 (JML, FB, \href{https://calclab.org}{https://calclab.org}), the ERC Consolidator Grant \textit{ProduSemy} (JML, Grant No. 101044282, see: \href{https://doi.org/10.3030/101044282}{https://doi.org/10.3030/101044282}). All claims expressed in this article are solely those of the authors and do not necessarily represent those of their affiliated organizations, or those of the publisher, the editors and the reviewers. Any product that may be evaluated in this article, or claim that may be made by its manufacturer, is not guaranteed or endorsed by the publisher. The funders have/had no role in study design, data collection and analysis, decision to publish or preparation of the manuscript.

\bibliography{custom}

\end{document}